\documentclass[letterpaper, 10 pt, conference]{ieeeconf}  

\IEEEoverridecommandlockouts                              

\pdfoutput=1
\usepackage{amsmath,amssymb,amsfonts}
\usepackage{xcolor}
\usepackage[]{graphicx}
\usepackage{wrapfig}
\usepackage{multirow}
\usepackage[caption=false,font=footnotesize]{subfig}
\title{\LARGE \bf
Learning Stable Normalizing-Flow Control for Robotic Manipulation
}

\author{Shahbaz Abdul Khader*$^{1,2}$, Hang Yin*$^{1}$, Pietro Falco$^{2}$ and Danica Kragic$^{1}$%
\thanks{Accepted for IEEE ICRA 2021. A supplementary video can be found at https://youtu.be/Vl8HFq-lk94}
\thanks{*These authors contributed equally (listed in alphabetical order).}
\thanks{$^{1}$Robotics, Perception, and Learning lab, Royal Institute of Technology, Sweden.
{\tt\small \{shahak, hyin, dani\}@kth.se}.}%
\thanks{$^{2}$ASEA Brown Boveri (ABB) Corporate Research, Sweden.
{\tt\small pietro.falco@se.abb.com}.}%
\thanks{This work was partially supported by the Wallenberg AI, Autonomous Systems and Software Program (WASP) funded by the Knut and Alice Wallenberg Foundation.}
}

\begin{document}

\maketitle
\thispagestyle{empty}
\pagestyle{empty}

\begin{abstract}

Reinforcement Learning (RL) of robotic manipulation skills, despite its impressive successes, stands to benefit from incorporating domain knowledge from control theory. One of the most important properties that is of interest is control stability. Ideally, one would like to achieve stability guarantees while staying within the framework of state-of-the-art deep RL algorithms. Such a solution does not exist in general, especially one that scales to complex manipulation tasks. We contribute towards closing this gap by introducing \textit{normalizing-flow} control structure, that can be deployed in any latest deep RL algorithms. While stable exploration is not guaranteed, our method is designed to ultimately produce deterministic controllers with provable stability. In addition to demonstrating our method on challenging contact-rich manipulation tasks, we also show that it is possible to achieve considerable exploration efficiency--reduced state space coverage and actuation efforts--without losing learning efficiency.

\end{abstract}

\section{INTRODUCTION}
\label{sct:intro}
Reinforcement Learning (RL) promises the possibility of robots to autonomously acquire complex manipulation skills. With the advent of deep learning, neural networks have been used as policies of a rich form that can even connect input, e.g. visual images, to output, e.g. motor torque or velocity, in an end-to-end manner~\cite{levine2016end, googlegrasping2018ijrr}. Although such general purpose application of RL is interesting, methods that benefit from prior domain knowledge, either as special structures for the policy or general properties from control theory, are gaining momentum. Employing structure on polices include trajectory-based shallow policy search such as ~\cite{theodorou2010generalized, stulp2012pathicml, hyin2014Hum} and also embedding variable impedance controller (VIC) structure into a deep neural policy~\cite{martin2019iros, ludovicvariable2020ral}. Such methods typically show a faster learning curve when compared to more general deep RL methods. What is more interesting is the trend to push the state-of-the-art by incorporating the property of control stability into RL as a means to guarantee safe and predictable behavior~\cite{perkinsandbarto2003jmlr,berkenkamp2017safe,controlbarrier_aaai2019,variancereduct_icml2019,stableRAL2020khader}. Unfortunately, with exception of one~\cite{stableRAL2020khader}, no such methods have been applied in the context of robotic manipulation. 

Stability is one of the first properties to be ensured whenever a closed-loop control law is synthesized. Despite this, stability guaranteed deep RL algorithms that actually scale to real robot manipulators have proven to be elusive. Part of the reason is that stability analysis assumes knowledge of the dynamics model, something that is unavailable in an RL problem. Therefore, an RL algorithm would either need to learn a dynamics model first and then update policy~\cite{berkenkamp2017safe}, assume prior knowledge of nominal or partial models in order to reason about stability~\cite{controlbarrier_aaai2019,variancereduct_icml2019}, or assume prior stabilizing controllers~\cite{perkinsandbarto2003jmlr,variancereduct_icml2019}. These methods seldom scale to manipulation control which often involves contact with the environment. Learning~\cite{modellearnRAL2020khader} or even offline modeling of contact dynamics is notoriously difficult. A recent advancement was made that exploited the passive interaction property between manipulator and its environment that allowed to skip the difficult step of model learning of contact dynamics~\cite{stableRAL2020khader}. However, this work employed specialized policy structure and Cross Entropy Method (CEM) based constrained policy search that can only use one entire trajectory as one data point; whereas, state-of-the-art policy gradient methods can utilize all state-action samples as data points along with gradient information. Therefore, achieving stability in deep RL methods and its potential implications on learning efficiency remains an open question.

\begin{figure}[t]
\centering
\includegraphics[width=0.48\textwidth]{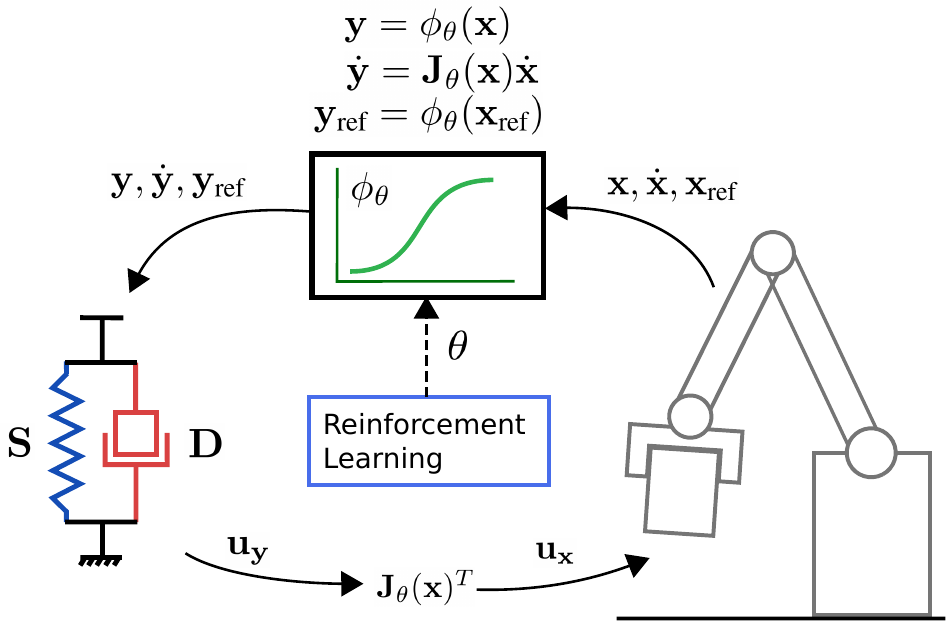}
\caption{\textbf{Learning Normalizing-flow Controllers} A 'normal' spring-damper system operating in a transformed space regulates the transformed image of the manipulation task dynamics. The transformation is through a bijection $\phi_{\theta}(.)$ (\textit{normalizing-flow}), with Jacobian $\mathbf{J}_{\theta}$, that is parameterized as a neural network and learned through reinforcement learning. The optimum $\phi(.)$ that is just enough for the fixed spring-damper system to solve the desired task is sought. Such a construction imparts the stability properties of the spring-damper system to the entire system. $\mathbf{x},\Dot{\mathbf{x}},\mathbf{x}_{\text{ref}}$ are the position, velocity and regulation point in the real system, and $\mathbf{y},\Dot{\mathbf{y}},\mathbf{y}_{\text{ref}}$ its counterparts in the transformed space. $\mathbf{u}_{\mathbf{x}}$ and $\mathbf{u}_{\mathbf{y}}$ are control action forces.}
\label{fig:normflow_summary}
\end{figure}

We present an approach to learn stable \textit{normalizing-flow} controllers for manipulation tasks. The focus is put on contact-rich tasks. Unlike the prior work~\cite{stableRAL2020khader}, we do not target stable exploration but aim for stability guaranteeing deterministic controllers as a final result of RL. The proposed method uses neural network parameterized policies with additional structure and is suitable for any state-of-the-art policy search method. Our results show that not only is learning stable controllers feasible, without sacrificing learning efficiency, but also that it is possible to achieve significant reduction in state space coverage and actuation efforts--both desirable in real-world RL. We reason that this improvement in exploration efficiency is due to the fact that the stability property, while acting as a bias to the policy, is able to direct the exploration towards the goal. The results are supported with extensive simulation as well as a real-world robotic experiments on contact-rich manipulation tasks.


\section{Preliminaries}
\label{sct:preliminaries}
\subsection{Reinforcement Learning}
\label{sct:preliminaries:rl}
Reinforcement learning addresses solving a Markov decision process problem that is described by an environment model (or dynamics) $p(\mathbf{s}_{t+1}|\mathbf{s}_{t},\mathbf{a}_{t})$, a state space $\mathbf{s}\in\mathcal{S}$, an action space $\mathbf{a}\in\mathcal{A}$ and a reward function $r(\mathbf{s}_t, \mathbf{a}_t)$. The environment is generally assumed to be unknown. The solution to the problem is reached when the policy $\pi_{\theta}(\mathbf{a}_t|\mathbf{s}_t)$ of the agent, that acts upon the environment, is able to maximize the accumulated reward, usually along the horizon of a fixed length $T$: 
\begin{equation*}
\label{eqn:rl_opt}
    \theta^* = \underset{\theta}{\operatorname{argmax}}\  \mathbb{E}_{\mathbf{s}_{0}, \mathbf{a}_{0}, ..., \mathbf{s}_{T}}[\sum_{t=0}^Tr(\mathbf{s}_t, \mathbf{a}_t)],
\end{equation*}
where $\{\mathbf{s}_{0}, \mathbf{a}_{0}, ..., \mathbf{s}_{T}\}$ is a sample trajectory from the distribution induced in the stochastic system.

Although on-policy RL methods typically require a stochastic policy, continuous control problems such as manipulator control usually expects a deterministic policy as the ultimate result. With this in mind, we reserve the term controller exclusively for a deterministic map of state to action, $\mathbf{a}=\pi(\mathbf{s})$, and the term policy for the stochastic counterpart, $\mathbf{a}\sim\pi(\mathbf{a}|\mathbf{s})$. The quantity $\pi$, therefore, is interpreted based on context. We shall use the notation $\mathbf{u}$ also for action to maintain conformity to control literature.

\subsection{Robot Manipulator Dynamics}
\label{sct:preliminaries:robotdyn}
The state of a robot manipulator is defined by a generalized coordinate $\mathbf{x} \in \mathbb{R}^{d}$, and its time derivative $\Dot{\mathbf{x}}$ ($\mathbf{s}=[\mathbf{x}^T\ \dot{\mathbf{x}}^T]^T$). The robot dynamics can be written as:

\begin{equation}
\label{equ:manipsys}
    \mathbf{M}(\mathbf{x})\Ddot{\mathbf{x}} + \mathbf{C}(\mathbf{x}, \Dot{\mathbf{x}})\Dot{\mathbf{x}} + \mathbf{g}(\mathbf{x}) = \mathbf{u} + \mathbf{f}_{\text{ext}}
\end{equation}
with $\mathbf{M}(\mathbf{x}) \in \mathbf{R}^{d\times d}$, $\mathbf{C}(\mathbf{x}, \Ddot{\mathbf{x}}) \in \mathbf{R}^{d\times d}$ and $\mathbf{g}(\mathbf{x}) \in \mathbb{R}^{d}$ denoting the inertia matrix, Coriolis term and the gravitational force. The manipulator is actuated by control efforts $\mathbf{u} \in \mathbb{R}^{d}$ and subject to external generalized force $\mathbf{f}_{\text{ext}} \in \mathbb{R}^{d}$.  

In the context of RL, the coupled system of manipulator interacting with its environment is considered the environment of the RL problem. Equation (\ref{equ:manipsys}) represents all the involved quantities: the policy acts through $\mathbf{u}$, the manipulator part of the environment is represented by the LHS and the robot-environment interaction is accounted by $\mathbf{f}_{\text{ext}}$. In this paper, we further assume a passive environment for the robot, where the robot is the only actuated entity. This has implications for the stability property and will be explained in Sec. \ref{sct:approch:stable_nf_control}.

\subsection{Lyapunov Stability}
\label{sct:preliminaries:lyapunov}
Lyapunov stability considers the temporal behaviour of nonlinear control systems such as robotic manipulators \cite{slotine1991applied}. The system is stable if the state trajectories that start close enough are bounded around an equilibrium point, and is asymptotically stable if the trajectories eventually converge to the equilibrium point. Such properties are the first necessary steps towards a guarantee that the system does not diverge rapidly causing potential harm.  Lyapunov analysis proves stability by constructing a Lyapunov function $V(\mathbf{x},\dot{\mathbf{x}})$, a scalar function of the state, with the property that it decreases as the controlled system evolves. See \cite{slotine1991applied} for a formal definition.

In the manipulator case, stability implies that any trajectory that is realized by the controller is bounded in state space and converges to an equilibrium point. The equilibrium point is usually designed to be a unique task relevant goal position, for example the final position in an episodic task.


\subsection{Invertible Neural Networks and Normalizing-Flows}
\label{sct:preliminaries:normflownet}
In normalizing-flows~\cite{normflowsuervey2020pami}, complex probability distribution models are constructed by learning a sequence of invertible and differentiable neural network transformations that maps simple probability distributions into the required complex ones. The overall transformation is a bijection that maps from one domain to another.

\begin{figure}[htb]
\centering
\includegraphics[width=0.48\textwidth, trim=0cm 5.5cm 0cm 5cm, clip]{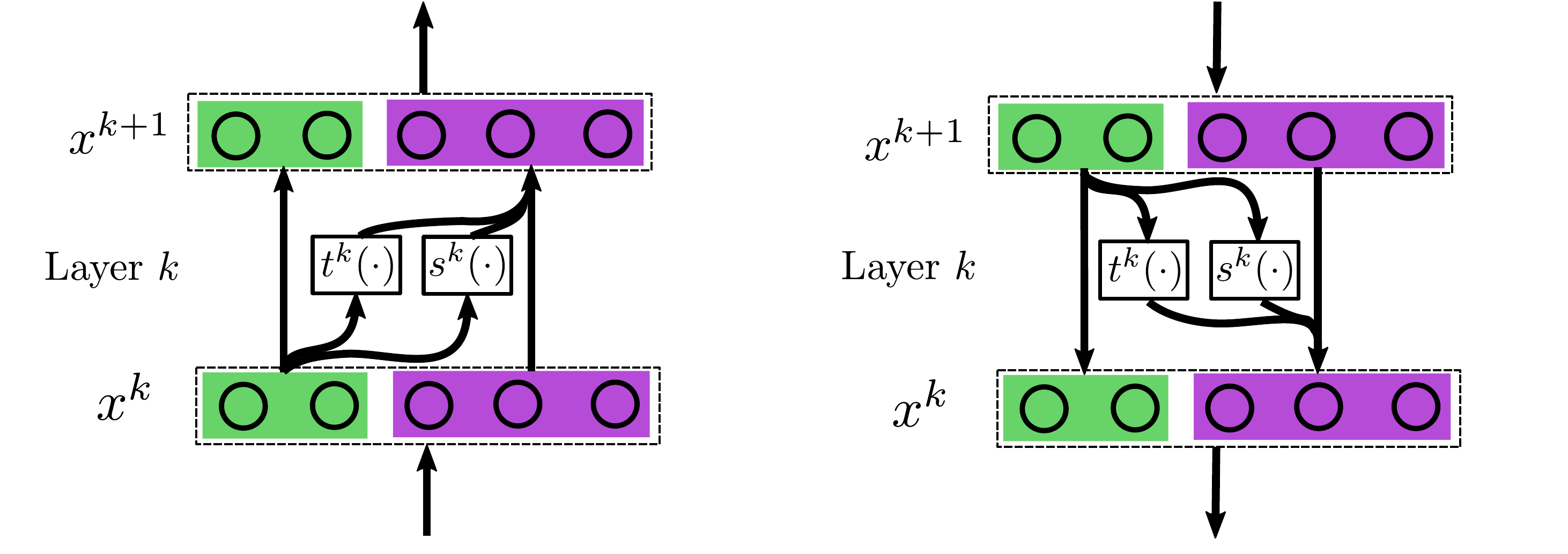}
\caption{\textbf{Normalizing-flow transformations} The layers can be stacked to build bijective neural network transformations.}
\label{fig:normflow_net}
\end{figure}

In a specific design, a normalizing-flow network was constructed by stacking invertible affine coupling layers~\cite{dinh+al-2017-density-iclr}:
\begin{equation*}
\label{equ:realnvp}
\begin{split}
    \mathbf{x}^{k+1} & = \phi^k(\mathbf{x}^k) \\
    \mathbf{x}^{k+1}_{1:d^k} & = \mathbf{x}^{k}_{1:d^k} \\
    \mathbf{x}^{k+1}_{d^k+1:d} & = \mathbf{x}^{k}_{d^k+1:d} \odot \text{exp}(\mathbf{s}^{k}(\mathbf{x}^{k}_{1:d^k})) + \mathbf{t}^{k}(\mathbf{x}^{k}_{1:d^k}) 
\end{split}
\end{equation*}
where $\mathbf{x}^k$ is the input to the layer, $D$ is the dimension of the input and $d^k < D$ is the index where the input dimensions are partitioned at layer $k$. The notation $\odot$ denotes elementwise multiplication and $\mathbf{s}^{k}(\cdot)$ and $\mathbf{t}^{k}(\cdot)$ are nonlinear transformations. See Fig.~\ref{fig:normflow_net} for an illustration. The entire network is composed as $\phi = \phi^{k}\circ \phi^{k-1}\circ...\circ\phi^{1}$ with the invertibility cascaded. The parameterization of the network boils down to $\mathbf{s}^{k}(\cdot)$ and $\mathbf{t}^{k}(\cdot)$. Fully-connected networks are an option. The nonlinear activation needs to be differentiable, such as the case in~\cite{rana2020euclideanizing}.

\section{Stable Normalizing-Flow Policy}
\label{sct:approch}
To realize the goal of RL with stability properties, we expect the policy to be: composed of a stable controller, differentiable, parameterized as a neural network for the most parts and amenable to unconstrained policy search. 

In this section, we first draw some parallels between density estimation and dynamical systems to motivate the notion of normalizing-flow controller. This is followed by the formal stability proof and also formalizing the RL problem.

\subsection{Normalizing-Flow for Second-order Dynamical Systems}
\label{sct:approch:nf_ds}
\begin{figure}[htb]
\centering
\includegraphics[width=0.48\textwidth, trim=1cm 7.5cm 0cm 7.5cm, clip]{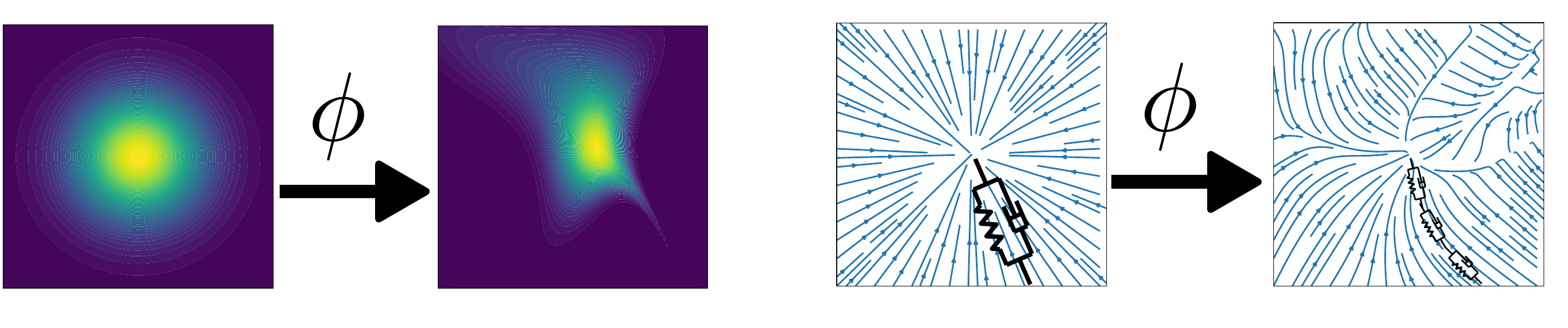}
\caption{\textbf{Left}: modeling complex probability distribution with normalizing-flow, e.g. by transforming from a simple normal distribution. \textbf{Right}: modeling complex attractor fields through a similar process of nonlinear transformation from 'normal' spring-damper system (e.g. $\mathbf{S}=\mathbf{D}=\mathbf{I}$).}
\label{fig:pdfandflowdyn}
\end{figure}

The main idea is transforming a simple regulation control system with desirable properties, such as a spring-damper system with stability, to a more complex one without losing those properties. This is analogous to normalizing-flow models that use bijective mappings to construct rich probability densities from a normal distribution, see Fig.~\ref{fig:pdfandflowdyn}. While these models transform random variables back and forth, our model is designed to transform the state variables. In both cases, the bijective transformations are learned from data. Our method is inspired from~\cite{rana2020euclideanizing}, but unlike it, we use a second-order spring-damper system and learn the transformation using RL.

Given a position space $\mathbf{x}\in\mathcal{X}$, a transformed position space $\mathbf{y}\in\mathcal{Y}$ and a nonlinear bijection $\mathbf{y}=\phi(\mathbf{x})$, consider the spring-damper regulation control:
\begin{equation*}
    \begin{split}
    \mathbf{u}_{\mathbf{y}} = -\mathbf{S}\mathbf{y}-\mathbf{D}\Dot{\mathbf{y}} \implies
    \mathbf{u}_{\mathbf{y}} = -\mathbf{S}\phi(\mathbf{x}) -  \mathbf{D}\mathbf{J}(\mathbf{x})\Dot{\mathbf{x}}
    \end{split}
\end{equation*}
where $\mathbf{J}$ is the Jacobian of $\phi$. From the principle of virtual work, we expect control in the $\mathbf{x}$ space $\mathbf{u}_\mathbf{x}$ to yield $\mathbf{u}_\mathbf{x}^T \delta\textbf{x} = \mathbf{u}_\mathbf{y}^T \delta\textbf{y}$. Using $\delta\mathbf{y} = \mathbf{J}(\mathbf{x})\delta\mathbf{x}$, we have:
\begin{equation}
\label{equ:nfctrl}
\mathbf{u}_{\mathbf{x}} = \mathbf{J}^T(\mathbf{x})\mathbf{u}_{\mathbf{y}} = -\mathbf{J}^T(\mathbf{x})\mathbf{S}\phi(\mathbf{x}) -  \mathbf{J}^T(\mathbf{x})\mathbf{D}\mathbf{J}(\mathbf{x})\Dot{\mathbf{x}}
\end{equation}

We call the form in Eq.~\eqref{equ:nfctrl} a \textit{normalizing-flow controller}. Note that the transformation is comparable to general coordinate transformation in task space.

\subsection{Stable Normalizing-Flow Controller}
\label{sct:approch:stable_nf_control}
We present our formal theoretical result of the stable controller as following:

\textit{Theorem}: Given a differentiable and bijective function $\phi: \mathbb{R}^{d} \rightarrow \mathbb{R}^{d}$ with its Jacobian $\mathbf{J}: \mathbb{R}^{d} \rightarrow \mathbb{R}^{d \times d}$, positive definite matrices $\mathbf{S} \succ 0$ and $\mathbf{D} \succ 0$, the robot dynamics Eq. \eqref{equ:manipsys} driven by the controller
\begin{equation}
\label{equ:controller}
    \mathbf{u} = -\mathbf{J}(\mathbf{x})^T\mathbf{S}[\phi(\mathbf{x})-\phi(\mathbf{x_{\text{ref}}})] - \mathbf{J}^T(\mathbf{x})\mathbf{D}\mathbf{J}(\mathbf{x})\Dot{\mathbf{x}}
\end{equation}
is 1) globally asymptotically stable at the unique attractor $\mathbf{x_{\text{ref}}}$ if $\mathbf{f}_{\text{ext}}=\mathbf{0}$; and 2) a passive dynamical system if $\mathbf{f}_{\text{ext}} \neq \mathbf{0}$. We assume gravity is compensated externally.

The above theorem can be proved by performing a Lyapunov analysis with a function: 
\begin{equation*}
    V(\mathbf{x}, \Dot{\mathbf{x}}) = \frac{1}{2} [\phi(\mathbf{x})-\phi(\mathbf{x_{\text{ref}}})]^T\mathbf{S}[\phi(\mathbf{x})-\phi(\mathbf{x_{\text{ref}}})] + \frac{1}{2}\Dot{\mathbf{x}}^T\mathbf{M}(\mathbf{x})\Dot{\mathbf{x}}
\end{equation*}
Given positive definiteness of $\mathbf{S}$ and $\mathbf{M}(\mathbf{x})$, we have $V \geq 0$. Note that $\phi(\cdot)$ is bijective so the equality holds only when $\mathbf{x} = \mathbf{x_{\text{ref}}}$. By taking the time-derivative of the Lyapunov function, we obtain:
\begin{equation*}
    \Dot{V} = [\mathbf{J}(\mathbf{x})\Dot{\mathbf{x}}]^T\mathbf{S}[\phi(\mathbf{x})-\phi(\mathbf{x_{\text{ref}}})] + \Dot{\mathbf{x}}^T\mathbf{M}(\mathbf{x})\Ddot{\mathbf{x}} + \frac{1}{2}\Dot{\mathbf{x}}^T\Dot{\mathbf{M}}(\mathbf{x})\Dot{\mathbf{x}}
\end{equation*}
Substitute $\mathbf{M}(\mathbf{x})\Ddot{\mathbf{x}}$ with Eq. \eqref{equ:manipsys}, \eqref{equ:controller} and $\mathbf{f}_{\text{ext}}=\mathbf{0}$:
\begin{equation*}
    \begin{split}
    \Dot{V} & = \Dot{\mathbf{x}}^T[\mathbf{J}(\mathbf{x})^T\mathbf{S}(\phi(\mathbf{x})-\phi(\mathbf{x_{\text{ref}}})) + \mathbf{u} \\
    & + \frac{1}{2}(\Dot{\mathbf{M}}(\mathbf{x})-2\mathbf{C}(\mathbf{x}, \Dot{\mathbf{x}})) \Dot{\mathbf{x}}]    \\
            & = - \Dot{\mathbf{x}}^T\mathbf{J}^T(\mathbf{x}) \mathbf{D}\mathbf{J}(\mathbf{x})\Dot{\mathbf{x}} < 0 
    \end{split}
\end{equation*}
where the skew-symmetric property of $\Dot{\mathbf{M}}-2\mathbf{C}$ and $\mathbf{D} \succ 0$ is used. From Eq. \eqref{equ:controller} the only invariant set for $\Dot{V} = 0$ is $\mathbf{x} = \mathbf{x_{\text{ref}}}$ and $\Dot{\mathbf{x}} = \mathbf{0}$. This is because Jacobian $\mathbf{J}(\mathbf{x})$ of bijective $\phi(\cdot)$ is full rank at any $\mathbf{x}$. According to LaSalle's invariant set theorem~\cite{salle1961stability}, $\mathbf{x}$ tends to $\mathbf{x_{\text{ref}}}$ when $\Dot{V} = 0$ and the controlled dynamical system is globally asymptotically stable at $\mathbf{x_{\text{ref}}}$. 

By inserting $\mathbf{f}_{\text{ext}}$ to the equation above and observing that
\begin{equation*}
    \begin{split}
    \Dot{V} & = \Dot{\mathbf{x}}^T\mathbf{f}_{\text{ext}} - \Dot{\mathbf{x}}^T\mathbf{J}^T(\mathbf{x}) \mathbf{D}\mathbf{J}(\mathbf{x})\Dot{\mathbf{x}} <  \Dot{\mathbf{x}}^T\mathbf{f}_{\text{ext}},
    \end{split}
\end{equation*}

we conclude that the controlled dynamical system is passive under $\mathbf{f}_{\text{ext}}$. The significance of passivity is that when two stable and passive systems interact physically, the coupled system is also stable~\cite{hogan2018impedance,colgate1988robust}. In other words, by ensuring the robot to be passive in isolation, we can conclude that all physical interaction with unknown but passive environments are automatically stable. This property can be exploited to avoid modeling or learning unknown interaction dynamics. See~\cite{stableRAL2020khader,khansari2014modeling} for examples. 

The above result is obtained for a generalized coordinate $\mathbf{x}$. In the case of Cartesian space of a redundant manipulator, where the position coordinate represented by $\mathbf{x}$ is not a generalized coordinate system, we propose to implement the Jacobian transpose control presented in Sections 8.6.2 and 9.2.2 of \cite{siciliano2016springer}. This also includes a slightly reformulated Lyapunov analysis that involves joint space kinetic energy term and joint space viscous friction. As explained in Section 8.6.2 of \cite{siciliano2016springer}, such a formulation guarantees energy decrease for any joint space velocity and thus ensures all null-space motion to eventually cease and reach an equilibrium position.  

\subsection{RL with Normalizing-Flow Controller}
\label{sct:approch:stable_nf_policy}
The controller in Eq.~\eqref{equ:controller} is an instance of the deterministic controller $\mathbf{a}=\pi(\mathbf{s})$ and can be easily converted into its stochastic counterpart $\pi(\mathbf{a}|\mathbf{s})$ by incorporating an additive Gaussian noise. Furthermore, by implementing $\phi(\mathbf{x})$ as a normalizing-flow neural network according to Sec. \ref{sct:preliminaries:normflownet}, all the variable terms in Eq.~\eqref{equ:controller} including $\mathbf{J}(\mathbf{x})$ are differentiable. Thus, it is only required to drop our policy into any policy gradient model-free RL algorithm such as Proximal Policy Optimization (PPO)~\cite{ppo2017shulman}.

It is important to note that only the controller has stability guarantee and not the noise-incorporated policy. We expect the RL algorithm to eventually refine the noise covariance to low values such that the resulting controller after RL is good enough for the task. This is a reasonable expectation for continuous control tasks such as manipulator control. Regardless of obtaining a stable controller, we would also like to investigate if the additional structure of our policy will compromise the learning efficiency. This is because the stability property that acts as an attractor toward the goal position ($\mathbf{x_{\text{ref}}}$) can be seen as a bias to an otherwise unbiased policy. It is important to establish whether this bias is helpful during the learning process or not.


\section{Experimental Results}
\begin{figure}[t]
 \vspace{2mm}
    \centering
    \subfloat
       {\includegraphics[width=0.35\textwidth]{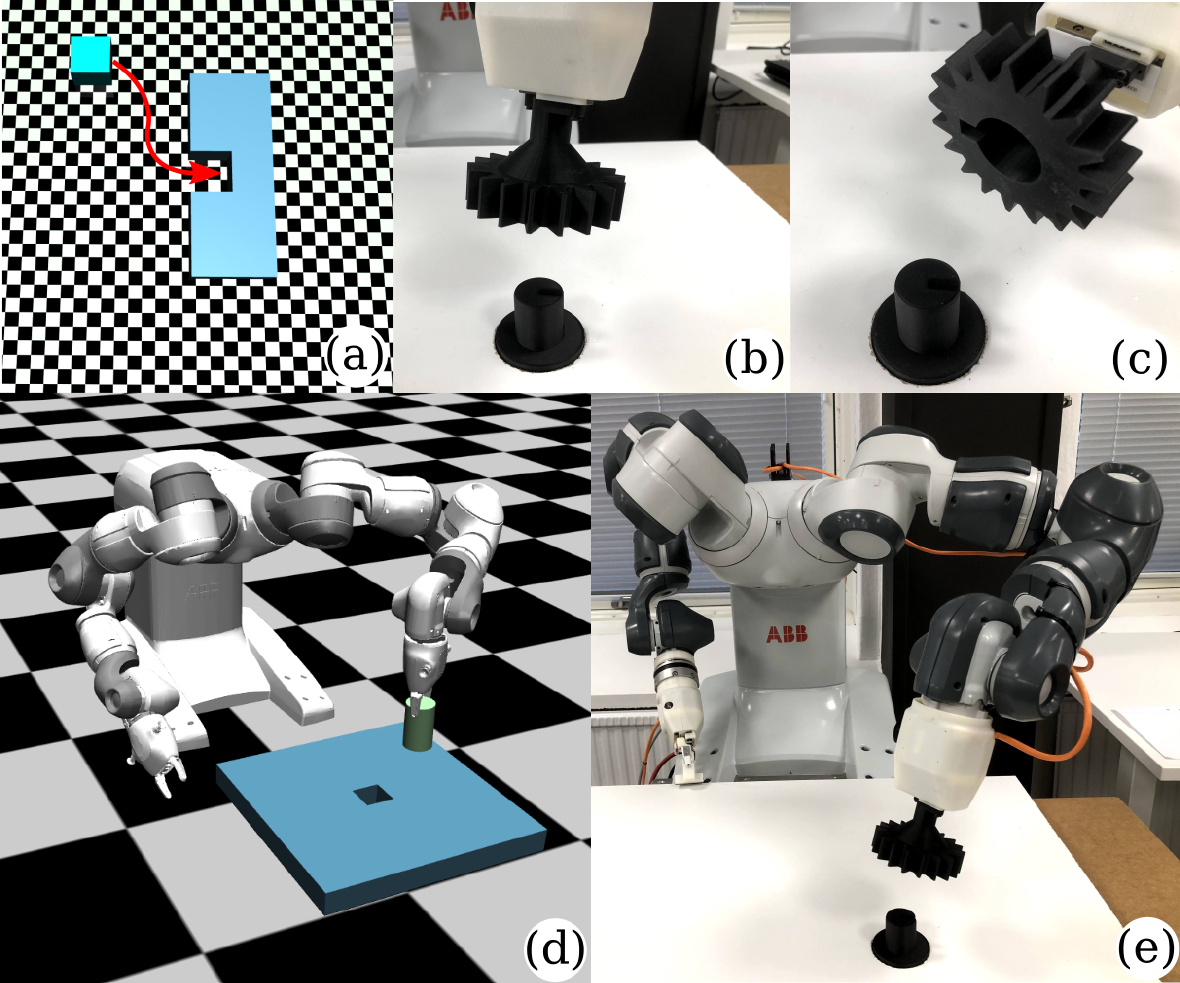}}\\
     \caption{\textbf{Experimental setup} \textbf{(a)} 2D Block-insertion: the block has dimensions of $50\times50\times50$ mm, weight of $1$ Kg and insertion clearance of $2$ mm. An example path is indicated from a representative initial position. \textbf{(b, c \& e)} Real-world gear assembly: a gear with a cylindrical hole of diameter $30$ mm and depth $20$ mm is to be assembled onto a cylindrical shaft of $28$ mm diameter. \textbf{(d)} Simulated peg-in-hole: a cylindrical peg of $23$ mm diameter is to be inserted into a square hole of width $25$ mm and depth $20$ mm. An insertion depth of $25$ mm for block-insertion and $16$ mm for the robot assembly tasks are considered successful.}
     \label{fig:exp_setup}
\end{figure}

The proposed method is validated on a number of tasks with varying dimensionality and complexity. While the low-dimensional block-insertion task (Fig. \ref{fig:exp_setup}a) provides a ground for better visualization of the basic principles involved, the simulated peg-in-hole (Fig. \ref{fig:exp_setup}d) and the real-world gear assembly (Fig. \ref{fig:exp_setup}e) tasks help validate manipulation tasks that we care about. All the three tasks are episodic tasks with fixed goal positions. Furthermore, they are also contact-rich applications that provide the additional challenge of learning the force interaction behaviour--not just the motion profile--that suits well the second-order setting of force based policies.

Important hyperparameters and general experiment details are described below. The number of flow elements is indicated by $n_{\text{flow}}$, where a flow element is defined as two successive normalizing-flow layers (Sec. \ref{sct:preliminaries:normflownet}) with swapped partitioning of input dimensions. This is to avoid one partition always passing the identity mapping, as suggested in~\cite{dinh+al-2017-density-iclr}. The number of units in the hidden layer of $\mathbf{s}^k(.)$ and $\mathbf{t}^k(.)$, both of which are identically defined as fully-connected networks with only one hidden layer, is denoted as $n_h$. As mentioned earlier, the normalizing-flow controller is converted to a stochastic policy by incorporating an additive normal noise, whose value is initialized with a spherical standard deviation defined by the scalar $\sigma_{\text{init}}$. The stiffness and damping matrices, $\mathbf{S}$ and $\mathbf{D}$, are desired to be kept as identity matrices as per our interpretation of 'normal' spring-damper system. We use the deep RL framework \textit{garage}~\cite{garage} along with its default implementation of PPO. Note that $\sigma_{\text{init}}$ is only relevant for a specific case where the standard deviation of the policy is directly implemented as trainable parameters and not as a neural network. The reader is directed to the software implementation of \textit{garage} for more details. We used the reward function suggested in~\cite{levine2015learning} for all our experiments.

 \begin{figure}[t]
    \centering
    \subfloat
       {\includegraphics[width=0.49\textwidth]{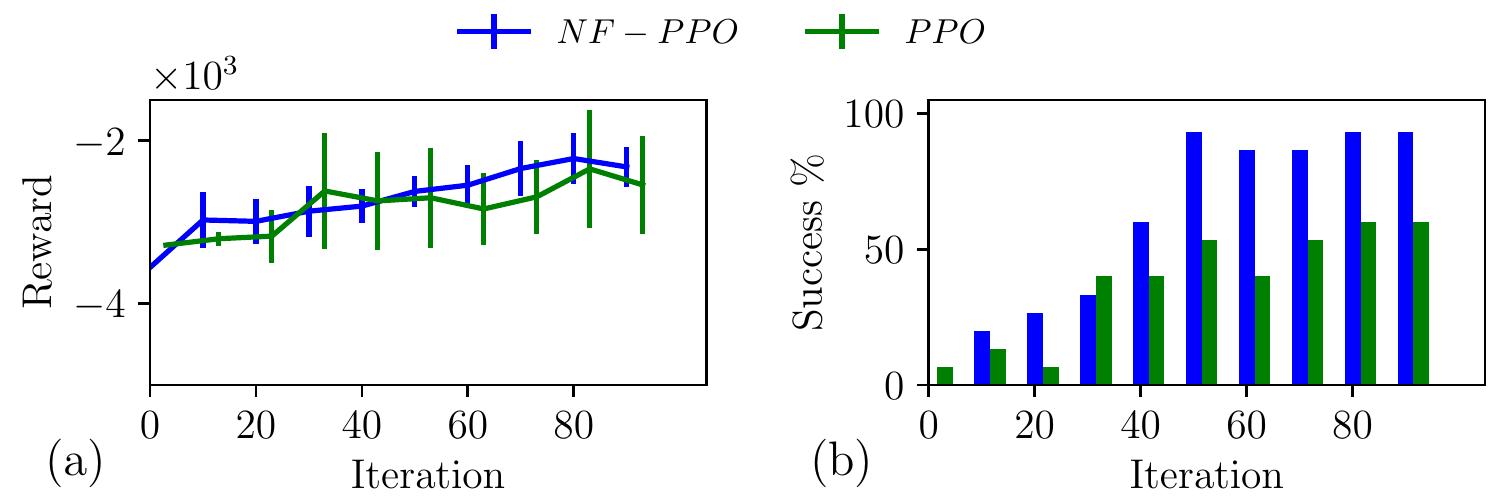}}\\
     \subfloat{\includegraphics[width=0.49\textwidth]{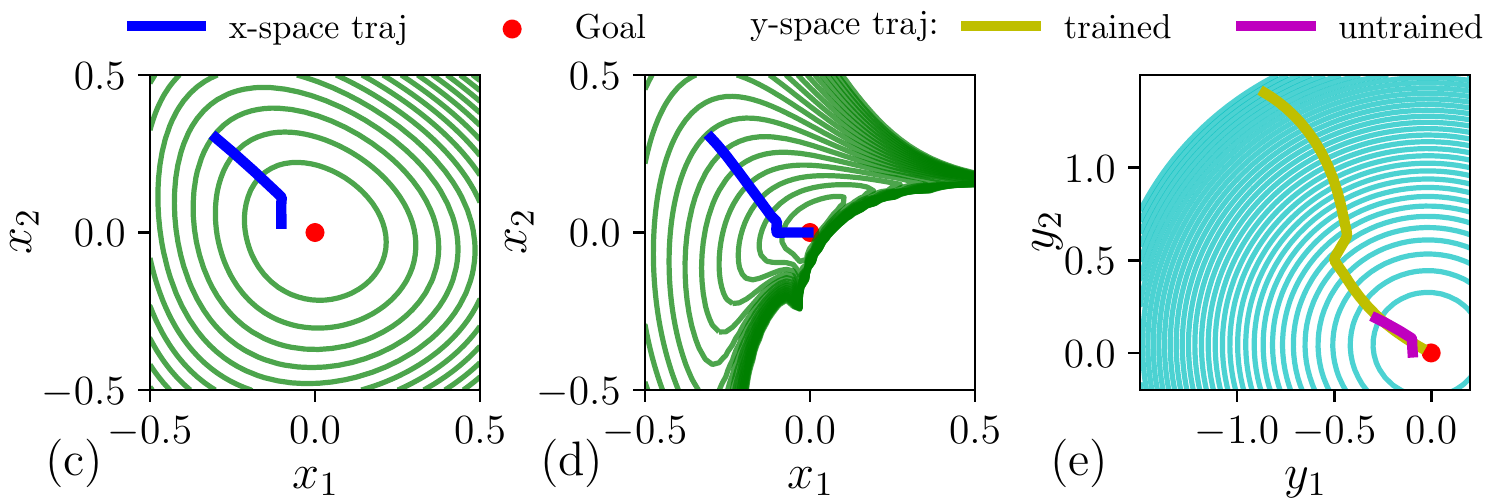}}
     \caption{\textbf{Learning Normalizing-flow policies for 2D Block-insertion} \textbf{(a)} RL reward \textbf{(b)} Success rate of insertions \textbf{(c)} A deterministic rollout, overlaid on the energy function $V_{|\dot{\mathbf{x}}=0}$ contour plot, in $\mathbf{x}$ space for randomly initialized $\mathbf{s}^k(.)$ and $\mathbf{t}^k(.)$ (iteration 0). \textbf{(d)} The same as \textbf{c}, but after RL (after iteration 99) \textbf{(e)} Plots corresponding to \textbf{c} and \textbf{d}, but in $\mathbf{y}$ space.}
     \label{fig:block2D_results}
\end{figure}

\subsection{2D Block-Insertion}
The 2D block-insertion task was first introduced in \cite{stableRAL2020khader} as a simplified representation of contact-rich manipulation. A block of mass, that can only slide on a surface without being able to rotate, is controlled by a pair of orthogonal forces acting on it with the goal of inserting it into a slot structure. Naturally, the 2D position $\mathbf{x}$ and velocity $\dot{\mathbf{x}}$ together forms the state variable and the 2D force $\mathbf{u}$ is the action variable. The final goal position, when inserted into the slot, is the regulation reference $\mathbf{x_{\text{ref}}}$. The initial position $\mathbf{x}_0$ is sampled from a wide normal distribution with the elements of its diagonal standard deviations as $50$ mm and $100$ mm, corresponding to the horizontal and vertical axes, respectively. We used 15 rollouts, each with an episode length of 2 seconds, in one iteration of RL. The remaining hyperparameters were set as $n_{\text{flow}}=2$, $n_h=8$, $\sigma_{init}=2.0$~N, and $\mathbf{S}=\mathbf{D}=\mathbf{I}$. 

In our $100$ iterations long experiment, we compare the RL performance of the proposed policy parameterization and also a regular neural network policy. The former will be hereafter called normalizing-flow PPO (NF-PPO) and the latter just PPO. Note that it is only the policy parameterization that is different while the RL algorithms in both cases are PPO. In Fig. \ref{fig:block2D_results}a, it can be seen that NF-PPO does have an advantage in learning efficiency as it achieves near $100\%$ insertion success rate at about 50 iterations. Figures \ref{fig:block2D_results}c-d show a single trajectory rollout of the deterministic controller before and after RL. The trajectory may be compared to the expected one in Fig. \ref{fig:exp_setup}a. The trajectory is overlaid on the contour plot of the proposed Lyapunov energy function ($V_{|\dot{\mathbf{x}}=0}$). When $\dot{\mathbf{x}}=0$, the policy is the negative gradient of the energy function, which can therefore be used to partially visualize the policy. It is interesting to note that even with randomly initialized $\mathbf{s}^k(.)$ and $\mathbf{t}^k(.)$, the policy is well-behaved and acts as an attractor field toward the goal. The situation is similar in Fig. \ref{fig:block2D_results}d except that the policy (and $V$) has been reshaped by RL into a richer one that not only behaves as an attractor but also accomplishes the task. Figure \ref{fig:block2D_results}e tracks the trajectories before and after RL in the transformed $\mathbf{y}$ space. In this space, the trajectories may be severely deformed by the learned $\phi(.)$, but the energy function is a fixed independent quadratic function. 

\subsection{Simulated Peg-In-Hole}
 \begin{figure}[t]
 \vspace{2mm}
    \centering
    \subfloat
       {\includegraphics[width=0.49\textwidth]{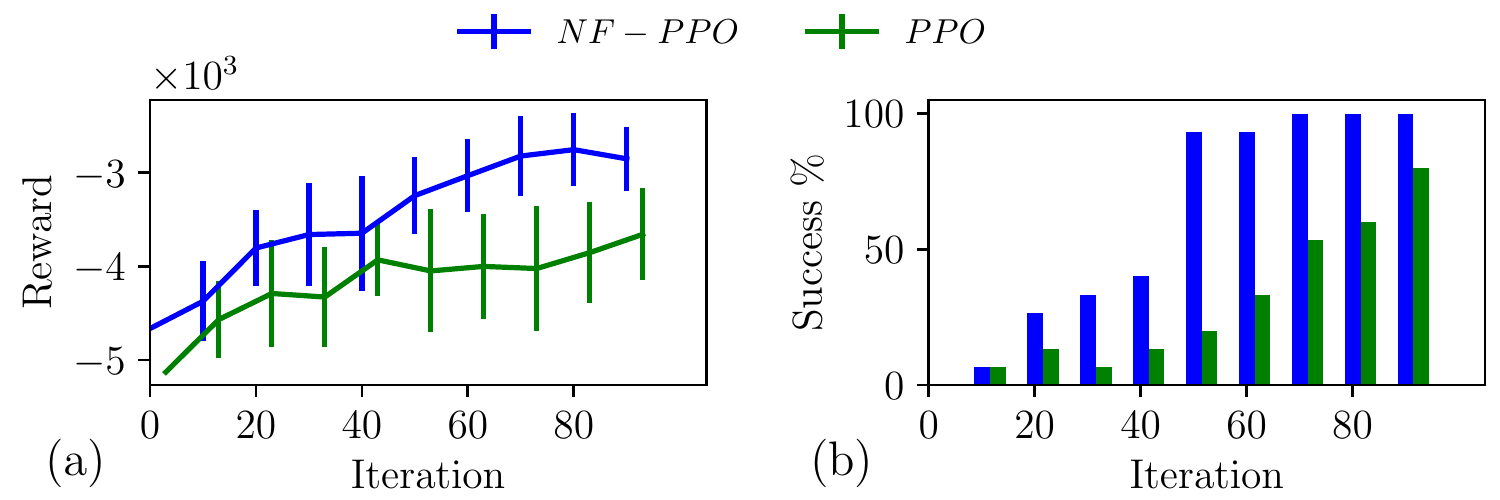}}\\
     \caption{\textbf{Learning Normalizing-flow policies for peg-in-hole} \textbf{(a)} RL reward \textbf{(b)} Success rate of insertions. Corresponds to $\sigma_{init}=1.0$ in Table \ref{tab:init_policy_var}.}
     \label{fig:peg_results}
\end{figure}

 \begin{figure}[t]
 \vspace{-4mm}
    \centering
    \subfloat[Iteration 0\label{fig:traj_samples:itr0}]{%
       \includegraphics[width=0.23\textwidth, trim=0.3cm 0.5cm 0cm 0cm, clip]{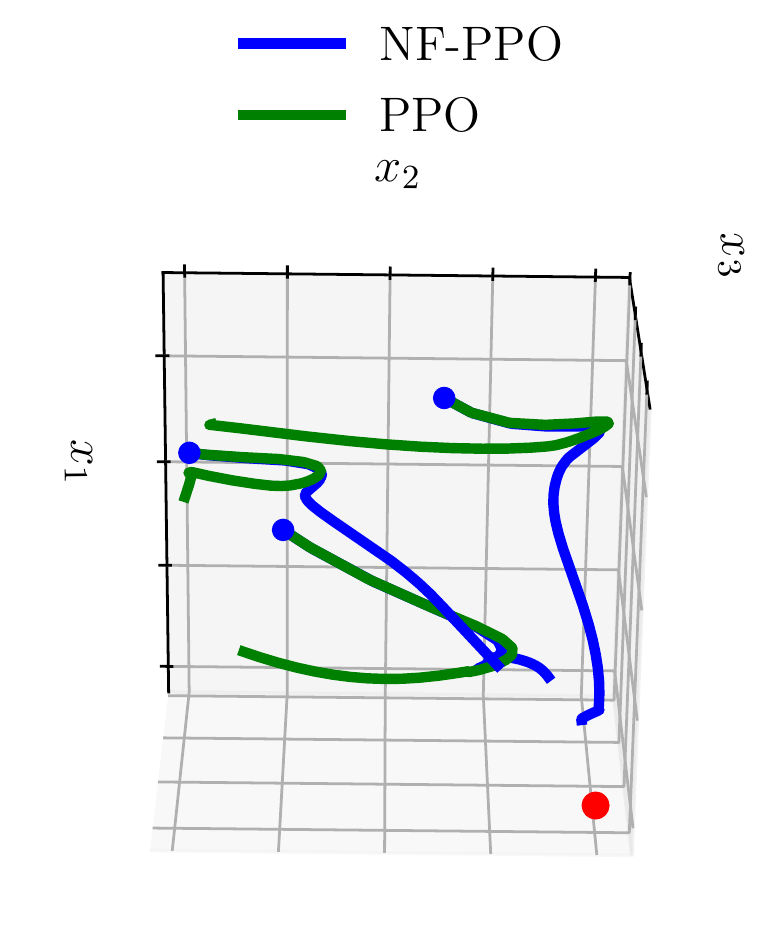}
     }%
     \subfloat[Iteration 9\label{fig:traj_samples:itr9}]{%
      \includegraphics[width=0.23\textwidth, trim=0.3cm 0.5cm 0cm 0cm, clip]{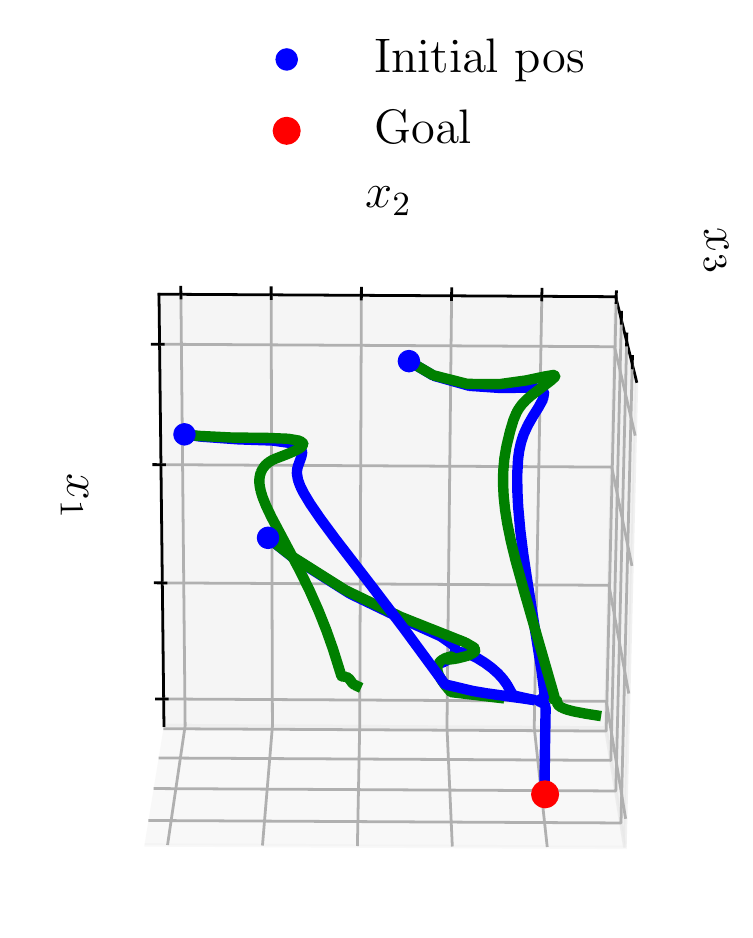}
     }\\
     \vspace{-2mm}
      \subfloat[Iteration 99\label{fig:traj_samples:itr99}]{%
      \includegraphics[width=0.23\textwidth, trim=0.3cm 0.5cm 0cm 0cm, clip]{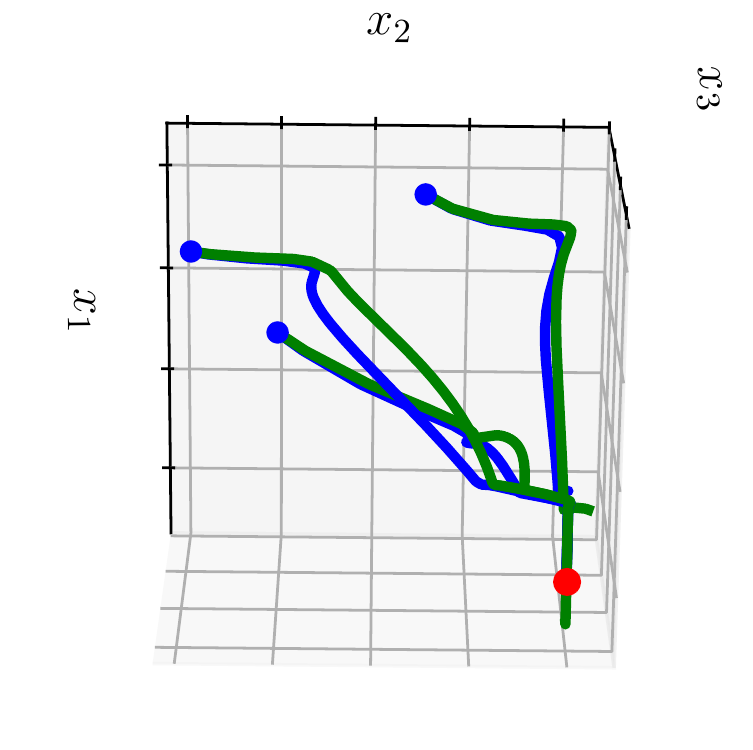}
     }%
      \subfloat[End-effector position\label{fig:traj_samples:state_dist}]{%
      \includegraphics[width=0.23\textwidth, trim=0cm 0cm 0cm 0cm, clip]{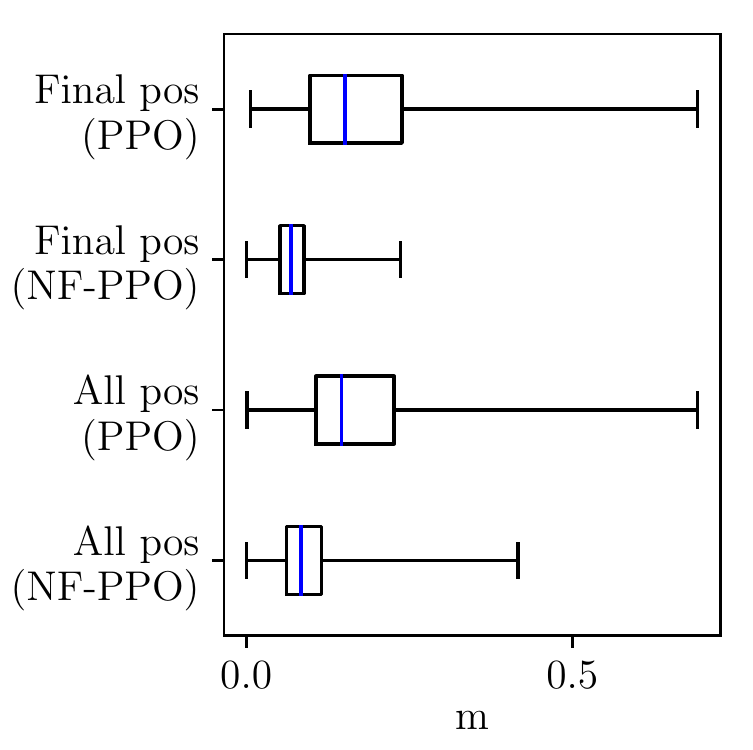}
     }
     \caption{\textbf{Effect of stable policy on learning peg-in-hole} \textbf{(a-c)} Deterministic rollouts of three randomly chosen initial positions at different stages of RL show consistent convergent behavior for NF-PPO. Note that these rollouts are only for evaluation and otherwise the policy is stochastic. \textbf{(d)} Distributions of end-effector distance $||\mathbf{x}-\mathbf{x_{ref}}||$ during the first 10 iterations of RL. The whiskers include all samples.
     }
     \label{fig:traj_samples}
\end{figure}

\begin{table}
\vspace{2mm}
 \caption{Impact of Initializing Policy Variance}
 \label{tab:init_policy_var}
\centering
\begin{tabular}{ |c|c|c|c|c| } 
 \hline
 &$\sigma_{\text{init}}$  & $\textit{Itr}_{90}$ & $||\mathbf{x}-\mathbf{x_{ref}}||$ & $||\tau||$ \\ 
 \hline
 Units &N  &  & m & Nm \\ 
  \hline
 \multirow{4}{*}{NF-PPO}& \textbf{1.0} & \textbf{63} & $\mathbf{0.1\pm0.05}$ & $\mathbf{0.9\pm1.1}$ \\ 
 &1.5 & 67 & $0.1\pm0.05$ & $1.4\pm1.1$ \\
 &2.0 & 60 & $0.12\pm0.1$ & $2\pm5.1$ \\
 &3.0 & 56 & $0.14\pm0.1$ & $2.7\pm2.1$ \\
 \hline
 \multirow{4}{*}{PPO}&1.0 & - & $0.17\pm0.1$ & $0.9\pm1.1$\\
 &1.5 & 85 & $0.18\pm0.1$  & $1.3\pm1.2$ \\ 
 &2.0 & 73 & $0.18\pm0.12$ & $1.8\pm3.4$ \\ 
 &\textbf{3.0} & \textbf{57} & $\mathbf{0.2\pm0.13}$ & $\mathbf{2.6\pm1.7}$ \\ 
 \hline
\end{tabular}
\end{table}

In this experiment, we scale up to a 7-DOF manipulator but only consider the translation part of the operational space for policy implementation and RL. The rotation part is implemented as a fixed regulation control (impedance control) that achieves the desired orientation for insertion. Our method can be extended to include rotation also but is left out for the sake of simplicity. The task itself is the peg-in-hole problem that has long been considered as a representative problem for contact-rich manipulation~\cite{nuttin1997learning,yun2008compliant,lee2019making}. The translation space position $\mathbf{x}$ and velocity $\dot{\mathbf{x}}$ form the state variable and the corresponding force $\mathbf{u}$ is the action variable. The initial position is sampled from a wide normal distribution in the joint space and then transformed into the operational space. The mean of the normal distribution roughly matches with the position in Fig. \ref{fig:exp_setup}d and the spherical standard deviation is chosen to be $0.2$ radians. The goal position, which is the inserted position, is set as the reference $\mathbf{x_{\text{ref}}}$. As in the previous case, here also we used 15 rollouts per iteration with each rollout having an episode time of 2 seconds. The remaining hyperparameters were set as $n_{\text{flow}}=2$, $n_h=16$, $\sigma_{init}=1.0$~N, and $\mathbf{S}=\mathbf{D}=\mathbf{I}$.

From the results of a $100$ iterations long RL experiment in Fig. \ref{fig:peg_results}a-b, it can be seen that there is a clear advantage for our method in terms of learning efficiency, although PPO eventually catches towards the end of $100$ iterations. To examine the effect of stability, we track deterministic rollouts from three randomly chosen initial positions at various stages of RL in Fig. \ref{fig:traj_samples}a-c. At iteration 0, with randomly initialized policy, PPO causes unpredictable motion away from the goal while NF-PPO ensures convergence towards the goal. The situation has improved for PPO in iteration 9 but still not on par with NF-PPO. Finally, at iteration 99, PPO appears to have attained similar behavior as NF-PPO, although PPO only managed 2/3 insertions compared to 3/3 by NF-PPO. Apart from predictable motion behaviors, another advantage we expect from the stability property is efficient exploration that is more directed towards the goal. Such a trend is visible in Fig. \ref{fig:traj_samples}d that presents the distributions of the end-effector distance relative to the goal during the first 10 RL iterations.

So far, we have not addressed the interplay between exploration and stability. In Table \ref{tab:init_policy_var}, we present results of extensive simulation study that reveals the impact of initial policy variance $\sigma_{\text{init}}$. It turned out that $\sigma_{\text{init}}$, a hyperparameter, drastically affects the overall exploration behaviour and hence RL performance. Note that the initial policy variance is a trainable parameter and the RL algorithm is expected to refine this to smaller values progressively. As $\sigma_{\text{init}}$ is increased, both NF-PPO and PPO show improvements in RL performance, indicated by the number of iterations required for consistent $90\%$ success ($Itr_{90}$). This gain is, of course, at the expense of generally undesirable conditions such as wider state distribution and also higher joint torques. Although exploration is necessary for RL, the real-world manipulator case imposes constraints due to safety concerns. Joint torques can be considered as a proxy for acceleration/deceleration and also contact forces, both of which should be kept to a minimum. Interestingly, the general drop in state space coverage of NF-PPO is not accompanied by a reduced joint torque. This indicates that exploration is as energetic as PPO but more efficiently directed. A more specific and significant result is that the learning performance of our method with $\sigma_{init}=1.0$ is closely matched by PPO only when $\sigma_{init}=3.0$. In other words, our method is exploration efficient and as a result reduces state space coverage by half and joint torques by a third. 

\subsection{Real-World Gear Assembly}

 \begin{figure}[t]
 \vspace{2mm}
    \centering
    \subfloat
       {\includegraphics[width=0.49\textwidth]{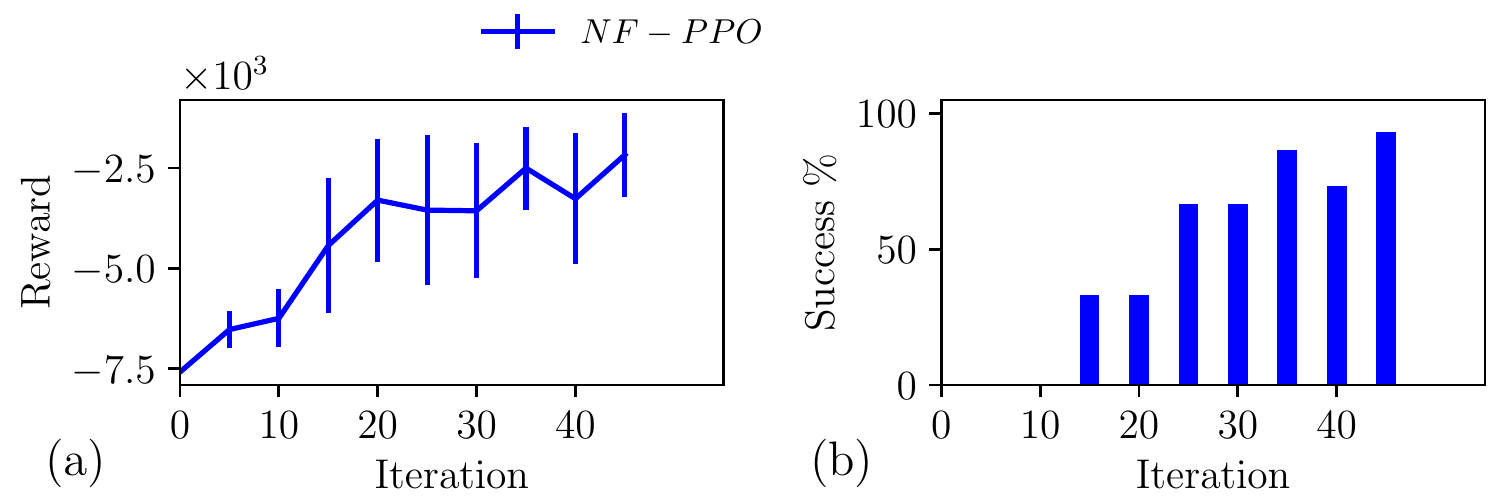}}\\
     \caption{\textbf{Learning Normalizing-flow policy for gear assembly} \textbf{(a)} RL reward. \textbf{(b)} Success rate of insertions.}
     \label{fig:gear_results}
\end{figure}
In this experiment, we successfully attempted a real robot experiment on the gear assembly task. This task is in essence the inversion of the peg-in-hole case, where the peg and hole are reversed. All aspects of the system are identical to the peg-in-hole case except that initial position is fixed, episode length is set to five seconds, and the following changes in hyperparameters are made: $n_{\text{flow}}=1$, $n_h=16$, $\sigma_{init}=5.0$~N, $\mathbf{S}=4.0\mathbf{I}$ and $\mathbf{D}=\mathbf{I}$. $\sigma_{init}$ and $\mathbf{S}$ are increased significantly to overcome uncompensated joint friction. The non identity value for $\mathbf{S}$ is against our interpretation of 'normal' spring-damper system but was chosen for expediting real-world RL, which would otherwise be time consuming.

Figure \ref{fig:gear_results} shows that the task was learned with $90\%$ success achieved at around 50 iterations. This is not directly comparable with the peg-in-hole simulation results in Fig. \ref{fig:peg_results} since the initial position is randomly chosen there. Regardless, the results are satisfactory and not entirely different from the simulation results. As in the case of simulated peg-in-hole, stable behaviour was apparent right from the beginning of the RL process. We did not attempt PPO on the real robot as it is not obvious how to safeguard against the generally unpredictable behaviour at the beginning of RL.

\section{Discussions and Conclusions}
In this paper, we sought out to investigate whether control stability can be incorporated into deep RL, and if so, what benefits would it entail? The approach was limited to formulating provably stable deterministic policies without ensuring stable exploration. Experimental evaluation focused on contact-rich tasks so that the force interaction aspect of manipulation is also included. Our results reveal that in addition to yielding a stable controller, the process of RL itself benefits from the approach.

Two concrete benefits can be identified for our method. First, the deterministic policy, which is usually the ultimate goal of RL, is provably stable for any possible instantiation of the policy, including randomly initialized or partially trained ones. This also includes the scenario where a fully trained policy visits a previously unseen region in the state space. In all such cases, trajectories will converge towards the goal providing a great deal of predictability and safety. Recall that stability is guaranteed even during contact with an unknown but passive environment. Second, the stability property has the potential to achieve efficient exploration. During random exploration, where stability is not technically guaranteed, the trajectories can still be directed towards the goal resulting in reduced state space coverage without affecting learning performance.

The most interesting outcome in our study is the concept of efficient exploration. Our results confirm that larger the exploration, the better it is in general for RL performance. However, this benefit comes with the generally undesirable cost of large coverage in state space and also high joint torques. Our results strongly indicate that by inducing stable behavior, state-action distribution can be drastically reduced without compromising the learning performance. This may be rephrased as: efficiently directing the exploration towards the general direction of the goal position can achieve considerable exploration efficiency. But, this advantage is lost when the exploration level is high enough to undermine the convergent behavior of stability, a possible explanation for the case $\sigma_{init}=3.0$ in Table \ref{tab:init_policy_var}. 

The proposed method is not without limitations. The policy structure in Eq. (\ref{equ:nfctrl}) is likely to be constrained in its modeling capacity compared to a completely free neural network. Unfortunately, no formal notion of model capacity exists and quantitative comparison cannot be readily made. Another noteworthy point is the higher training and inference time of our method. Recall that the policy contains additional structures including the Jacobian of the bijection, which is expensive to compute during inference time. This forced us to downgrade $n_{\text{flow}}$ to 1 from a value of 2 to meet the realtime requirements in the real-world experiment. This can be easily addressed by more efficient implementation and faster computing hardware. The presence of Jacobian in the policy also means that it has to be further differentiated for computing the policy gradient during training. Based on our experience, a fourfold increase in overall computation time can be expected when compared to standard deep policies.

The results in this work may be considered as one step closer towards provable control stability in the context of deep RL of robotic manipulation. We showed that learning stable deterministic policies, or controllers, is both feasible and useful through a number of simulated and real-world RL of manipulation skills. As a next step, we intend to pursue deep RL methods that feature stable exploration as well.





\bibliographystyle{IEEEtran}
\bibliography{references/rl,references/rl_skill,references/rl_skill_compliant,references/other,references/imitation_learning,references/control_opt_robotics,references/model_learning,references/ml,references/for_this_doc}

\end{document}